\documentclass[acmsmall,screen]{acmart}

\usepackage{amsmath,amsfonts}
\usepackage{algorithm}
\usepackage{algorithmic}
\usepackage{adjustbox}
\usepackage{graphicx}
\usepackage{textcomp}
\usepackage{xcolor}
\usepackage{xurl} 
\usepackage{tabularx}
\usepackage{tcolorbox}
\usepackage{listings}
\usepackage{mdframed}
\usepackage{multirow}

\definecolor{codegreen}{rgb}{0,0.6,0}
\definecolor{codegray}{rgb}{0.5,0.5,0.5}
\definecolor{codepurple}{rgb}{0.58,0,0.82}
\definecolor{backcolour}{rgb}{1.0,1.0,1.0}
\definecolor{backcolour_t}{rgb}{0.96,0.98,1.0}

\lstdefinestyle{mystyle}{
  backgroundcolor=\color{backcolour}, commentstyle=\color{codegreen},
  keywordstyle=\color{magenta},
  numberstyle=\tiny\color{codegray},
  stringstyle=\color{codepurple},
  basicstyle=\ttfamily\footnotesize,
  breakatwhitespace=false,         
  breaklines=true,                 
  captionpos=b,                    
  keepspaces=true,                 
  numbers=left,                    
  numbersep=5pt,                  
  showspaces=false,                
  showstringspaces=false,
  showtabs=false,                  
  tabsize=2
}

\AtBeginDocument{%
  }

\setcopyright{acmlicensed}
\copyrightyear{2025}
\acmYear{2025}

\begin{document}

\title{Integrating Large Language Models and Reinforcement Learning for Non-Linear Reasoning}


\author{Yoav Alon}
\affiliation{%
  \institution{University of Bristol}
  \country{United Kingdom}}
\email{yoav.alon@bristol.ac.u}

\author{Cristina David}
\affiliation{%
  \institution{University of Bristol}
  \country{United Kingdom}}
\email{yoav.alon@bristol.ac.u}

\renewcommand{\shortauthors}{Alon et al.}

\begin{abstract}
Large Language Models (LLMs) were shown to struggle with long-term planning, which may be caused by the limited way in which they explore the space of possible solutions.
We propose an architecture where a Reinforcement Learning (RL) Agent guides an LLM's space exploration:
(1) the Agent has access to domain-specific information, and can therefore make decisions about the quality of candidate solutions based on specific and relevant metrics, which were not explicitly considered by the LLM's training objective; (2) the LLM can focus on generating immediate next steps, without the need for long-term planning.
We allow non-linear reasoning by exploring alternative paths and backtracking.
We evaluate this architecture on the program equivalence task, and compare it against Chain of Thought \cite{DBLP:conf/nips/Wei0SBIXCLZ22} (CoT) and Tree of Thoughts~\cite{DBLP:conf/nips/YaoYZS00N23} (ToT). We assess both the downstream task, denoting the binary classification, and the intermediate reasoning steps.
Our approach compares positively against CoT and ToT.
\end{abstract}




\received{1 September 2024}

\maketitle

\section{Introduction}

Large Language Models (LLMs) have recently been successfully applied to a wide range of tasks, including code related tasks \cite{choudhuri2023far,PanICSE24,DBLP:journals/corr/abs-2402-09664,CodeRL}. When looking at the search strategies that they employ, LLMs are still mostly restricted to linear search during inference. For instance, the popular Chain of Thought (CoT) approach \cite{DBLP:conf/nips/Wei0SBIXCLZ22} attempts to break the reasoning process into smaller steps, denoted by intermediate thoughts of the LLM, thus enabling the LLM to perform logical reasoning. While multiple thought candidates corresponding to different paths through the search space exist at each step, the LLM always commits to one option, linearly building a solution. Several works have shown that this space exploration strategy may lead to difficulties with long-term proof planning \cite{DBLP:conf/iclr/Saparov023,DBLP:journals/corr/abs-2303-12712}. In particular, when multiple valid deduction steps are available, LLMs are not able to systematically explore the different options. 

Given the very large search space exhibited by many problems, exploration of different paths and backtracking may be beneficial. This problem was recently addressed by an approach coined Tree of Thoughts (ToT) \cite{DBLP:conf/nips/YaoYZS00N23}, which allows LLMs to make decisions by considering multiple different reasoning paths and self-evaluating choices to decide the next action.
While ToT was shown to enhance LLM’s problem-solving abilities for some tasks, it wasn't investigated in the context of code related tasks. Especially for code generation, the
LLM's ability to self-evaluate is particularly challenging due to the mismatch between training and inference metrics, which was highlighted by several works \cite{DBLP:conf/nips/Le0GSH22,DBLP:journals/corr/abs-2306-11816}. While models are trained using a next-token prediction objective, which maximizes the likelihood of the next ground-truth token, the code generated at inference is usually evaluated on its ability to compile and pass available unit tests. 

One objective of this paper is to examine the reasoning abilities of CoT and ToT for a code-related task (which involves code generation). This poses a challenge, as evaluating the accuracy of intermediate reasoning steps is inherently difficult. Most prior studies assess reasoning abilities indirectly by measuring performance on downstream tasks, such as mathematical problem-solving~\cite{DBLP:journals/corr/abs-2110-14168,DBLP:journals/corr/abs-2209-00840}. The only work we are aware of that explicitly considers intermediate reasoning steps uses a natural language question-answering task, where examples are derived from a synthetic world model represented in first-order logic~\cite{DBLP:conf/iclr/Saparov023}. However, it's unclear how to apply this approach to code-related tasks.

Instead, we focus on the well-established problem of answering \emph{program equivalence} queries, long studied in programming languages (via symbolic reasoning)~\cite{DBLP:conf/ac/Pitts00,DBLP:journals/dagstuhl-reports/LahiriMSU18,DBLP:conf/sas/IoossAR14,DBLP:journals/stvr/GodlinS13}: given programs $A$ and $B$, we ask whether they are semantically equivalent, meaning that they have the same I/O behaviour. The reasoning required to prove equivalence typically involves identifying a sequence of semantics-preserving transformations that convert program $A$ into program $B$. This can be represented as a series of mutated programs $A, A_1, \ldots, A_n$, where each such program $A_i$ preserves the behavior of the source program $A$, and shows increasing syntactical similarity to the target $B$ compared to the previous step $A_{i-1}$, until finally reaching $B$ itself. For illustration, Figure~\ref{fig:example} provides two equivalent programs $A$ and $B$ (both solutions of a Codeforces problem), and Figure~\ref{fig:transformations} provides the sequence of semantics preserving transformations leading from $A$ to $B$.

Program equivalence is a good candidate for evaluating intermediate reasoning steps because, as we frame it, these steps involve code generation, and their accuracy can be evaluated in a clear, structured manner. Specifically, we assess the accuracy of intermediate steps across three dimensions: syntactical correctness, functional correctness (preserving the behavior of the source program $A$), and syntactical similarity to the target program $B$. When programs $A$ and $B$ are equivalent, we know that the final program in the sequence of transformations should match $B$.

In addition to assessing the reasoning processes of CoT and ToT, we introduce and evaluate a novel architecture that expands upon the non-linear solution exploration concept of ToT. However, instead of relying on the LLM for evaluating candidate reasoning steps, this task is delegated to a Reinforcement Learning (RL) Agent equipped with domain-specific knowledge. This approach aims to overcome the limitations observed in LLM reasoning with CoT and ToT by placing the RL Agent in charge of the critical task of proof planning.


Similar to ToT, a reasoning tree of intermediate reasoning steps is lazily built: at each step, given the current node, the LLM is queried to generate a number of $n$ subsequent candidate intermediate steps; the Agent uses domain-specific information to decide whether to continue exploring one of them, or to backtrack to the previous node; for the latter, an unexplored child of that node may be subsequently chosen for exploration by the Agent, or it may backtrack further.
This architecture aims to take advantage of the Agent and LLM's respective strengths:
\begin{itemize}
\item The Agent has access to domain-specific information, and can therefore make decisions about the quality of the candidates based on specific and relevant metrics, which were not explicitly considered by the LLM's training objective.
\item  The LLM can focus on generating immediate next steps, without the need for long-term planning.
\end{itemize}

For the program equivalence task, the RL Agent leverages domain-specific information by utilising a syntax checker, running unit tests, and applying code similarity metrics.


\paragraph{Key findings} Our experiments led to several key findings, which are discussed in detail in the rest of the paper:

(1) For the program equivalence task, where multiple intermediate reasoning steps are needed, CoT often struggles to maintain accuracy of these steps. In our setting, this means that the intermediate programs frequently fail to maintain the same behaviour as the source program, and that, for equivalent programs, the final program in the transformation sequence has limited similarity to the target program.
This aligns with prior research, which concludes that when multiple valid deduction paths exist, LLMs often fail to systematically explore the different options resulting in poor long term planning~\cite{DBLP:conf/iclr/Saparov023}.

(2) While ToT does better than CoT with respect to both maintaining the behaviour of the source program, as well as increasing similarity with the target program, the results are still poor, especially for the latter aspect. We hypothesise that this is caused by the fact that the LLM has to self-evaluate programs in order to decide the next action. However, this evaluation requires domain-specific information, which was not part of the LLM's training objective.

(3) The Agent architecture outperforms CoT and ToT in the evaluation of both the intermediate reasoning and the downstream task.



\begin{figure}
    \centering
    
\textbf{Program $A$: Problem 266B, solution 7604802}
\begin{lstlisting}[language=Python]
n,t=list(map(int,input().split()))
a=input()
k=0
for i in range(t):
    a=a.replace('BG','GB')
print(a)
\end{lstlisting}

\textbf{Program $B$: problem 266B, solution 13328325}
\begin{lstlisting}[language=Python]
n,t = list(map(int,input().split()))
s = input()
while t>0:
    s = s.replace("BG","GB")
    t -= 1
print(s)
\end{lstlisting}
\caption{Example of two equivalent programs $A$ and $B$ denoting two solutions for the Codeforces Problem 266B: The programming task requires simulating the dynamic rearrangement of a queue consisting of boys and girls in a school canteen. Initially ordered as they entered, the boys in the queue allow the girls directly behind them to move forward each second. The programmer must create an algorithm to predict the final order of the queue after a specified number of seconds based on the initial arrangement and the defined movement rules.}
\label{fig:example}
\end{figure}


\begin{figure}
\lstset{style=mystyle}
\textbf{Program $A$}
\begin{lstlisting}[language=Python]
n,t=list(map(int,input().split()))
a=input()
k=0
for i in range(t):
    a=a.replace('BG','GB')
print(a)
\end{lstlisting}

\textbf{Program $A_1$: variable renaming a to s}
\begin{lstlisting}[language=Python]
n,t=list(map(int,input().split()))
s=input()
k=0
for i in range(t):
    s=s.replace('BG','GB')
print(s)
\end{lstlisting}
\textbf{Program $A_2$: remove redundant $k=0$}
\begin{lstlisting}[language=Python, caption=]
n,t=list(map(int,input().split()))
s=input()
for i in range(t):
    s=s.replace('BG','GB')
print(s)
\end{lstlisting}
\textbf{Program $A_3$: for to while conversion}
\begin{lstlisting}[language=Python]
n,t=list(map(int,input().split()))
s=input()
while t>0:
    s=s.replace('BG','GB')
    t -=1
print(s)
\end{lstlisting}
\textbf{Program $A_4$ (same as program $B$): replacing quotes}
\begin{lstlisting}[language=Python]
n,t= list(map(int,input().split()))
s = input()
while t>0:
    s = s.replace("BG","GB")
    t -=1
print(s)
\end{lstlisting}
\caption{Sequence of semantics preserving transformation from program $A$ to program $B$}
\label{fig:transformations}
\end{figure}

\paragraph{Contributions} We make the following key contributions:
\begin{itemize}
\item We designed an RL and LLM cooperative approach for the exploration of the solution space, suitable for tasks that require long-term planning, and where the success depends on 
 domain-specific criteria that were not
explicitly considered during the LLM’s training. 
\item We evaluated our architecture on program equivalence queries, where we investigated both the downstream task, which is the classification result, and the intermediate proof steps. For this task, our architecture compared positively against CoT and ToT prompting.

\item To the best of our knowledge, we performed the first investigation of a non-linear space
exploration for a code-related task, and  showed that it improves results compared to linear solution
building.



\end{itemize}

\paragraph{Roadmap} We start by describing our RL Agent architecture in Sec.~\ref{sec:approach}, followed by evaluating it and comparing it with CoT and ToT in Sec~\ref{sec:exp} and Sec.~\ref{sec:results}. We then discuss threats to validity and related works in Sec.~\ref{sec:threats-validity} and Sec.\ref{sec:related-work}, respectively.

\section{Our Approach}
\label{sec:approach}

\begin{figure*}[ht]
    \centering
    \includegraphics[width=\textwidth]{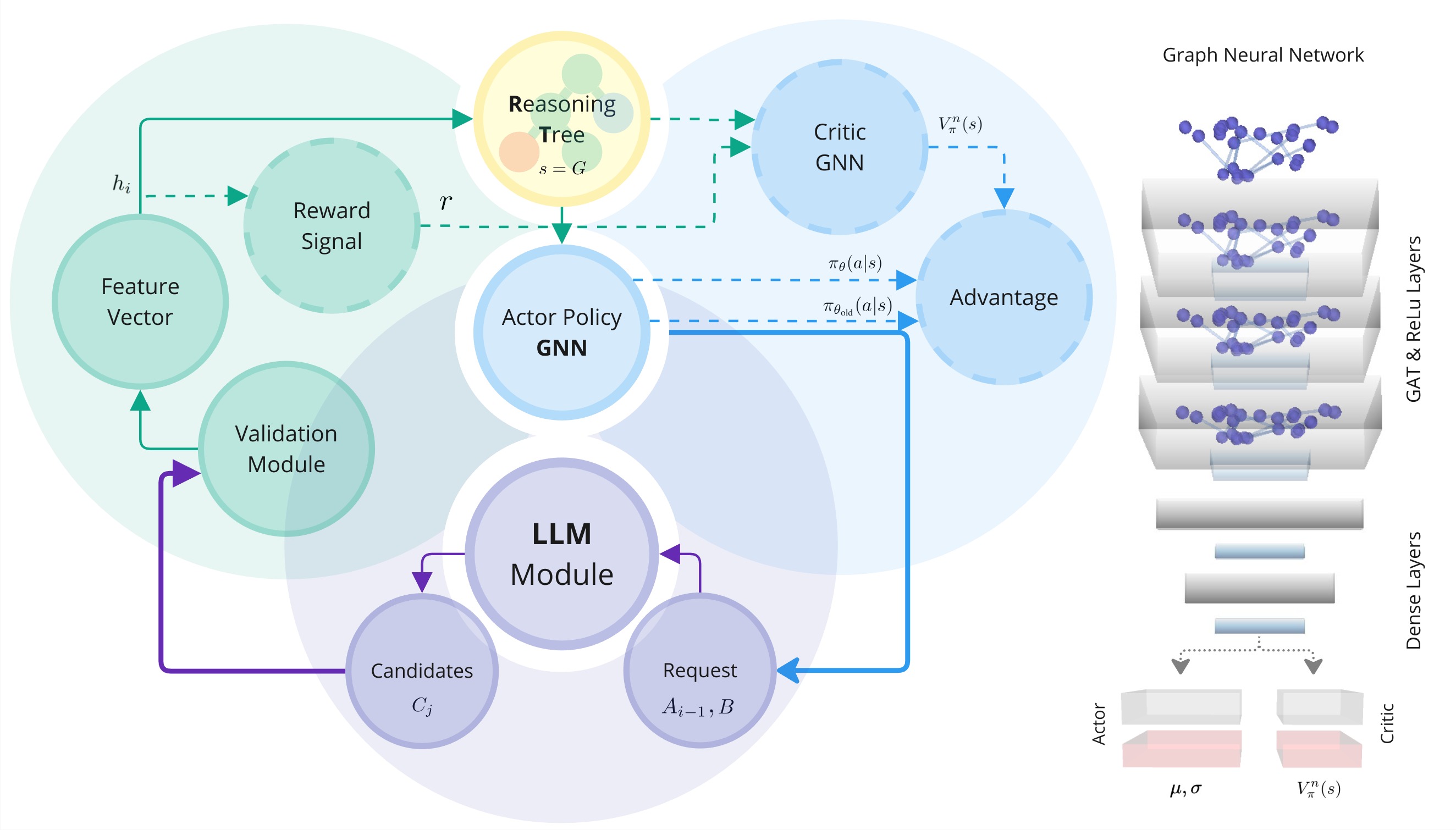}
    \caption{
    The proposed architecture (left) is composed of four components: the LLM Module (purple), the Validation Module (green),
    the Reasoning Tree (yellow), and the Agent (blue). The Graph Neural Network architecture (right) is used for actor and critic. (Best viewed in color). }
    \label{fig:architecture}
\end{figure*}




When solving a task, our approach generates the intermediate reasoning steps $A, A_1, \cdots A_n$ by lazily building a reasoning tree of such steps. For the equivalence task, the nodes in the reasoning tree are mutated programs (i.e. mutated versions of the source program $A$), with edges corresponding to the semantics-preserving transformations that lead from a parent program to a child program. The root of the reasoning tree is the source program $A$.
Examples of semantics-preserving transformations are variable renaming, changing the order of a function's arguments, modifications to control flow statements (e.g. switching the order of the branches of an \texttt{if} statement -- of course, with the corresponding negation of the guard, switching a \texttt{while} loop for a \texttt{for} one), substitution of data structures. 



The overall structure of our architecture consists of four components (See Figure~\ref{fig:architecture}), namely the LLM Module, the Validation Module, the Reasoning Tree and the RL Agent, which interact as described next:

At each step, 
given the current mutated program $A_i$, the {\bf LLM Module} queries the LLM, requesting $n$ candidates for the next mutated program $A_{i+1}$ obtained by applying a semantics-preserving transformation to $A_i$. 
Next is the {\bf Validation Module}, responsible for assessing 
the characteristics of each of the $n$ candidates. 
For each candidate, all the corresponding assessments are joined into a feature vector. The feature vectors are then added to the {\bf Reasoning Tree}, as the children of the currently explored node (the actual programs are stored separately).  
Lastly, the {\bf RL Agent} guides the exploration of the reasoning tree and decides which node to explore next. Namely, the RL Agent decides whether to continue exploring one of the currently generated children, or to backtrack to the previous node; for the latter, an unexplored child of that node may be subsequently chosen for exploration by the Agent, or it may backtrack further.

\subsection{LLM module}
\label{subsec:llmapi}
The LLM module is responsible for sending requests to the LLM and collecting the candidate mutated programs. Initially, the prompt includes the source program $A$ and the target program $B$. In subsequent iterations $i{>}1$, the prompt is based on the most recent mutated program $A_{i-1}$ and the target program $B$. Each time, we request $n$ candidates.

\subsection{Validation module}
\label{subsec:validation}

This module's objective is to take advantage of domain-specific information to compute metrics for the generated candidates, which were not explicitly part of the LLM's training objective. For the current problem, we are interested in syntactical correctness, functional correctness, code similarity, and granularity of the applied transformation.
Next, we explain how they are computed.


For \textbf{Syntactical Correctness} of a program, we check whether its Abstract Syntax Tree (AST) is structurally valid. The resulting output is either 1 (i.e., the program is syntactically correct) or 0 (i.e., the program is not syntactically correct).

We measure the \textbf{Functional Correctness} of a program as the percentage of unit tests that it successfully passes. This measures the code's functional correctness under the predefined test cases.

We use CodeBLEU  to assess \textbf{Code Similarity} 
    between two programs. 
    CodeBLEU is specifically designed for source code as it combines the strength of BLEU~\cite{DBLP:conf/acl/PapineniRWZ02} through the n-gram match with information about the code syntax via ASTs and code semantics via data flow. Intuitively, when computing the code similarity between a candidate and the target program, we want it to provide an indication of how many semantics-preserving transformations are still required to get from the candidate program to the target (as opposed to simply indicating the number of different syntactic tokens). Thus, we chose CodeBLEU as opposed to something like BLEU. CodeBLEU
    also shows a high correlation with
    human evaluation. It
    gives an output score between 0 and 100, where a CodeBLEU score of 100 means that the two programs perfectly match.

Given two programs, where the second was obtained from the first one by applying a transformation, \textbf{Granularity} is meant to estimate the coarseness of the said transformation.
    For this purpose, we use the Jaccard index \cite{codeSimilarityReview}, which compares the ASTs of the two programs by
    evaluating the overlap between their respective sets of nodes. The Jaccard coefficient takes values between 0 and 1, with 0 indicating no overlap and 1 complete overlap between the sets. For our use case, a larger value corresponds to a finer grained transformation.
    Granularity is of interest to us as we hypothesise 
    that smaller transformations would result in better reasoning abilities by breaking the overall task into small, manageable subtasks. This could also help the user to understand the transformation steps and gain trust in the solution.
    

Let's assume we have generated mutated programs $A_1, \cdots, A_{i-1}$, and we are now querying the LLM for program $A_i$. For each candidate $C_j, j=\overline{1,n}$ returned by the LLM, we compute:
 
\begin{description}
    \item[$\nu(C_j)$: ] Syntactical correctness of $C_j$. 
    \item[$\rho(C_j)$: ] Functional correctness of $C_j$. 
    \item[$\sigma(C_j, A)$: ] Code similarity to the source program $A$. 
    \item[$\sigma(C_j, A_{i-1}) $: ]  Code similarity to the last generated program $A_{i-1}$ (i.e., the parent node in the reasoning tree). This reflects changes over time.
    \item[$\sigma(C_j, B)$: ] Code similarity to the target program B. This serves as an indication of how many semantics-preserving transformations are still required to get from $C_j$ to $B$.
    \item[$g(C_j, A_{i-1})$: ] Granularity of the last applied transformation. 
    \item[$g(C_j, B)$: ] Jaccard index of the current candidate and the target program. This provides syntactic information in the form of the dissimilarity of the ASTs corresponding to the two programs.
\end{description}

All these metrics are then compiled into a feature vector $h_j$, which the Agent uses as a foundation for making decisions:
 \begin{equation}\label{eq:node_features}
 \begin{split}
 h_j & = [\nu(C_j), \rho(C_j), \sigma_(C_j, A), \sigma(C_j, A_{i-1}), \\
 & \quad \sigma(C_j, B), g(C_j, A_{i-1}), g(C_j, B)]
 \end{split}
 \end{equation}

In our experiments, we investigated the influence weights for each feature in a feature vector, and determined that the most important are syntactical and functional correctness, as well as similarity to the target program, which is very intuitive.

\subsection{Reasoning tree}
The reasoning tree $G$ is lazily built as explained next. At each step, given the current node corresponding to mutated program $A_{i}$ (initially corresponding to the source program $A$, which is the parent node of the reasoning tree), 
we generate $n$ candidates, $C_j, j=\overline{1,n}$. At this point, the reasoning tree $G$ is updated by adding an edge from the current node $A_{i}$ to each candidate $C_j$. 

Although, for simplicity, we may refer to the nodes in the reasoning tree as the mutated programs, these nodes only store the corresponding feature vectors, while the actual programs are stored separately.

\subsection{Agent module} 

The RL Agent's role is to determine the most effective sequence of semantics-preserving transformations of source program $A$ such that it becomes as similar as possible to target program $B$.  

\paragraph{Action space} The action performed by the Agent guides the traversal of the reasoning tree $G$. At each step, given the current node corresponding to mutated program $A_{i}$, and the newly generated $n$ candidates, $C_j, j=\overline{1,n}$,
the RL Agent can pick from $n + 1$ available actions: 
it can pick to explore one of the $C_j$ candidates or to backtrack to its immediate ancestor node, $A_{i-1}$.
If the latter option is picked, then, in the next iteration, the Agent will be able to pick to explore one of the still unexplored children of $A_{i-1}$ (i.e., a sibling of the current node) or further backtrack. Backtracking is disabled for the root node of the reasoning tree.

\paragraph{RL architecture} We make use of the actor-critic framework \cite{Sutton}, where an Agent (the ``actor'') learns a policy to make decisions, and a value function (the ``critic'') evaluates the actions taken by the actor by estimating their value or quality. The policy is denoted by $\pi_{\theta}(s=G,a)$  which represents the probability of taking action $a$ in state $s$. The value function (corresponding to the critic), denoted by $V(s)$, estimates the expected 
reward starting from state $s$. After training, at inference time, the actor Agent will use the learned policy to make decisions about the exploration of the reasoning tree.

In our setting, $G$ serves as the state $s$ for the RL framework. This is necessary in order to enable the possibility of backtracking, where the next action depends on the history of states (remember that, at each step, there is an option to jump back to the parent).

To effectively extract features from the current reasoning tree G, we employ a Graph Attention Network (GAT) \cite{GAT}, which is a variant of a Graph Neural Network (GNN) that leverages attention mechanisms. GATs have been empirically shown to outperform similar GNN models for applications involving code \cite{TerminationGNN}. 
At a high level, the method of feature extraction from the graph through graph convolution relies on two primary actions: first, gathering the characteristics of neighboring nodes for each individual node; second, updating each node using trainable weights and adjustable attention mechanisms. For more details, please consult \cite{GAT}.
As shown in Figure \ref{fig:architecture} (right), our model features three GAT layers, as referenced in \cite{GAT}. 
 The feature vectors in the initial GAT layer correspond to those computed by the Validation Module, denoted as $h_i^{(0)}$. The first hidden layer outputs features into a space $h_i^{(1)}$, with a dimensionality set to 30. Similarly, for the next layer. The critic's output dimension is set to 1 to provide a single estimate of the state value. Meanwhile, the actor policy's output has a dimensionality of 2, representing $\mu$ and $\sigma$ for a multivariate distribution. A normalized sample from this distribution is transformed into a discrete action within a space of $(n+1)$ options.

\paragraph{Training of the Agent}
The training process for the Agent begins with an initial random policy, $\pi_{\theta}(s=G,a)$. 
During training, we utilize experience replay, and the reward signal $r$ is subject to discounting, encouraging the actor policy to prioritize decisions that are optimal over the long term. Training is specifically conducted on equivalent programs and 
the reward signal is designed to encourage the creation of a sequence of semantics-preserving transformations from $A$ to $B$.
As criteria for long-term goals, we expect the semantics preserving transformations to eventually rewrite $A$ into $B$.
The discounted future reward, denoted as  $D_t$, is calculated as 
\begin{equation}
\label{eq:discount}
D_t = \sum_{k=0}^n \gamma^k R_{t+k+1}
\end{equation}
where $t$ denotes the number of steps until the Agent achieves the final goal, and  $\gamma$ is the discount factor, $0 < \gamma \leq 1$. 
For low values of $\gamma$, the Agent considers immediate rewards, while a gamma close to one assigns greater weights to future rewards. 
We continue training the Agent until the policy stabilizes and shows signs of convergence for the validation set. We employ transfer learning to adjust the Agent to a different base LLM, which uses significantly fewer training episodes than needed for complete training. 

As a note, we initially experimented with providing additional rewards for things like 
applying finely-grained transformations. However, it proved more effective to allow the policy to autonomously determine the importance of each available option.

\subsection{Termination Condition}

Defining a termination condition is crucial for the practical deployment of our algorithm. 
We stop the generation of new mutated programs when one of the following conditions is met:
\begin{itemize}
    \item  A maximum number of steps $m$ have been taken where the generated programs either didn't pass any of the unit tests or the similarity to the target program has not improved. This allows the quality of the generated programs to temporarily decrease/not improve.
    \item A maximum number of steps $p$ have been taken overall.
\end{itemize}    

\subsection{Adapting the architecture to a different task}

While we only applied the Agent architecture to the specific problem of answering program equivalence queries, we
believe it may be beneficial to other tasks that
require long-term proof planning (i.e. several intermediate reasoning steps), and where the success depends on criteria that were not explicitly considered during the LLM's training.

Adapting the framework would involve: defining the domain-specific evaluation criteria (e.g. in automated summarisation, the evaluation could involve metrics like summarisation ratio, factual consistency, and user engagement). Subsequently, the RL Agent needs to be trained  to optimise these criteria, while building and exploring a decision tree that captures the domain-specific features of each intermediate step.

\section{Experimental Evaluation}
\label{sec:exp}


\subsection{Experimental setup}

We run our experiments on a machine AMD Ryzen 9 6900HX Processor withNvidia GeForce RTX 3070 TI with 8GB RAM. All LLMs are accessed through APIs provided by third party services.

We developed a custom implementation of the policy-gradient algorithm in PyTorch, incorporating graph attention convolutions from the Deep Graph Library (DGL) for both the actor and critic. This approach enables us to effectively integrate graph-based data with RL frameworks.

For our experiments, we use $n=10$, meaning that we request ten candidates from the LLM. For the termination condition, we use $m=3$ and $p=10$.

The experiments are conducted twice, and the average metrics from both runs are then calculated and presented.


\subsection{Dataset}

We make use of a subset of the AlphaCode dataset \cite{AlphaCode_2016}, which consists of programming challenges of varying lengths and difficulties extracted from several coding challenge platforms. Each challenge includes a natural language description and unit tests, along with multiple solutions in Python 2 and C++ 4.3.2. In this work, we focus on Python, so we only use those solutions.  Moreover, we only picked those challenges that had at least 50 unit tests. The resulting dataset consists of 7,764 programming problems. 
%

The average number of lines of code and unique tokens (e.g., instruction, variable name) for each program significantly varies by difficulty level. For instance, easy challenges from CodeChef samples have an average of 15 lines of code and 144 unique tokens, while hard challenges average 40 lines of code and 204 unique tokens.

We reserve 75\% of the dataset for the training of the RL Agent. However, given that we only train until the performance on the validation set stabilises, we observe that the policy begins to stabilize as early as after 200 to 400 requests to the LLM (not episodes). Also, we employ transfer learning to adjust an Agent trained for a certain LLM to a different one,
which uses significantly fewer training episodes. 
%
Additionally, we keep 10\% of the dataset for validation purposes. That leaves us with 1164 programming problems for evaluation with a mean of 48.96 solutions per programming problem with an average of 26.08 lines of code and 110.71 unique tokens. For each problem, we take the combination of all the solutions to form pairs of equivalent programs.

In our investigation of intermediate reasoning steps (Sec.~\ref{sec:intermediate-evaluation}) and the additional Agent evaluation (Sec.~\ref{sec:agent-evaluation}), we focus exclusively on pairs of equivalent programs. This allows us to evaluate the reasoning process more effectively, as the objective is clear: apply a sequence of semantics-preserving transformations to the source program $A$ until it becomes the target program $B$. For the investigation of the downstream task (Sec.~\ref{sec:downstream-evaluation}), we also add pairs of non-equivalent programs. For this purpose, we mutate one program in all equivalent pairs, such that each pair is no longer equivalent -- typically through straightforward alterations like changing "+1" to "+2". 

\subsection{Considered LLMs}

In our evaluation, we consider the following LLMs: GPT-3.5 \cite{GPT-3.5}, GPT-4 \cite{GPT4}, and GPT-4 Turbo \cite{GPT-4-turbo}.
GPT-3.5, an earlier iteration in the GPT series, is known for its robust natural language processing abilities, though it occasionally struggles with nuanced context understanding. GPT-4 
offers more accurate and context-aware responses. GPT-4 Turbo, a streamlined variant of GPT-4, is designed for faster response times. 
%
We use the default values for the hyper-parameters of the LLMs (e.g. for GPT: Temperature: 1, Top\_p-nucleus sampling: 0.9, Frequency Penalty: 0, Presence Penalty: 0).

\subsection{Prompting}

For the Agent architecture, we experimented with several prompts and chose the following template: \textit{Given programs \{A\} and \{B\}, transform the first program so that it becomes syntactically more similar to the second program while retaining its semantics. Apply only one atomic transformation. Provide only the source code without your comments.} This prompt is used by the LLM module in Fig.~\ref{fig:architecture} to obtain candidate programs at each intermediate reasoning step. For the experiments in Sec.~\ref{sec:downstream-evaluation}, where we evaluate the downstream task given the generated intermediate reasoning, we provide the sequence of generated transformations in the prompt. 

For CoT, we used the prompt suggestion in ~\cite{DBLP:conf/nips/KojimaGRMI22}.
To trigger the ToT reasoning,
we used the ToT prompt proposed in \cite{tree-of-thought-prompting}: \textit{``Imagine n different experts are answering this question.
All experts will write down 1 step of their thinking,
then share it with the group.
Then all experts will go on to the next step, etc.
If any expert realizes they're wrong at any point, then they leave. The question is ...''}. 
We experimented with a few values for the number of experts, $n$, and we found that the optimal performance is achieved with three agents. When the number exceeds three, the LLM often groups agents together, leading to a severe decline in performance. We tested various LLM settings, such as adjusting the maximum output tokens, but did not manage to improve the results. 

\section{Experimental Results}
\label{sec:results}

We structure our results around three key areas. First, we evaluate and compare the intermediate reasoning steps generated for the CoT prompt, ToT prompt and the Agent architecture.  Next, we examine how these intermediate steps contribute to answering the downstream program equivalence queries. Finally, we conduct a deeper analysis of the Agent architecture to identify its most critical components and to assess whether a simpler architecture could achieve similar performance.
%

\subsection{Evaluation of the intermediate reasoning steps}
\label{sec:intermediate-evaluation}

In order to assess the intermediate steps taken by the Agent architecture, CoT prompting and ToT prompting, for each of them we measure:
Syntactical Correctness, Functional Correctness, Code Similarity to the target program, and Granularity where appropriate.
 For Syntactical and Functional Correctness, we are interested in both mean and final values: (1) \textbf{mean} represents the average metric calculated across all intermediate mutated programs within the sequence of mutated programs explored for each benchmark. For syntactical correctness we compute the percentage of programs that are syntactically correct; (2) \textbf{final} denotes the metric associated with the last mutated program generated for each benchmark. 
%
The numbers in the table are the average of mean and final values across all the benchmarks. The \textbf{final} values for Syntactical Correctness in the table denote the percentage of final programs that are syntactically correct.

For ToT, given that the experts
explore candidates concurrently, which they share with the rest after each step, it's difficult to determine which intermediate steps were actually useful and should be considered when computing the mean metrics. Thus, Table~\ref{tab:evaluation_Agent_ToT_final} only contains the values for the final mutated program (and no granularity results).

All the results used for this evaluation are captured in Table~\ref{tab:evaluation_Agent_CoT} and Table~\ref{tab:evaluation_Agent_ToT_final}. 

\paragraph{\textbf{RQ1(a). What are the intermediate reasoning capabilities of CoT and ToT prompting for the program equivalence task?}}

Both CoT and ToT provide poor quality intermediate reasoning steps. While they are generally able to generate intermediate programs that are syntactically correct, the functional correctness drops, meaning that the transformations frequently fail to maintain the semantics of the original program. Even more concerning, the similarity between the final mutated program and the target program $B$ is very low. This is particularly true for CoT, where it ranges between 46.05\% and 51.62\% for the different LLMs. ToT does a bit better with respect to code similarity, ranging between 60.52\% to 63.2\%.  

These findings support the hypothesis that CoT in particular has issues with long-term planning, which has been suggested by existing works for other problem domains~\cite{DBLP:conf/iclr/Saparov023} --
generating a sequence of semantically equivalent mutated programs that are progressively more similar to a target program requires long range planning.
For ToT, we hypothesise that its relatively poor reasoning is due to the fact that the intermediate candidates need to be assessed based on metrics that are different from the ones used during training. Namely, syntactical correctness, functional correctness and code similarity are different from the next-token prediction objective, which maximizes the
likelihood of the next ground-truth token.



\begin{table*}
\centering
\caption{Comparison between the Agent architecture and CoT with respect to the intermediate reasoning steps (best results in bold font).}
\label{tab:evaluation_Agent_CoT}
\begin{adjustbox}{width=1.0\textwidth, center}
\begin{tabular}{lcc|cc|c|c}
\hline
\textbf{Model} & \multicolumn{2}{c|}{\textbf{Syntactical Correctness (\%)}} & \multicolumn{2}{c|}{\textbf{Functional Correctness (\%)}} & \multicolumn{1}{c|}{\textbf{Code Similarity}} & \textbf{Granularity} \\
 & mean & final & mean & final & final & mean \\
\hline
GPT-3.5 \textbf{+ CoT}    & 75.96   & 76.92   & 31.90 & 34.62 
& 46.05 & 0.3879  \\
GPT-3.5 \textbf{+ Agent}  & \textbf{81.25}   & \textbf{78.4} & \textbf{72.34} & \textbf{52.15} 
& \textbf{87.15} & \textbf{0.4892} \\
\hline
GPT-4 \textbf{+ CoT}      & 97.06  & 97.06   & 65.08 & 61.03 
& 48.56 & 88.05  \\
GPT-4 \textbf{+ Agent}    & \textbf{98.75}  & \textbf{99.2} & \textbf{82.03} & \textbf{70.45} 
& \textbf{95.45} & \textbf{91.24} \\
\hline
GPT-4-turbo \textbf{+ CoT}      & 96.90  & 97.24   & 69.38 & 64.21 
& 51.62 & 86.44  \\
GPT-4-turbo \textbf{+ Agent}    & \textbf{98.54}  & \textbf{99.01} & \textbf{85.47} & \textbf{78.81} 
& \textbf{95.74} & \textbf{90.39} \\
\hline
\end{tabular}
\end{adjustbox}
\end{table*}

\begin{table*}
\centering
\caption{Comparison between the Agent architecture and ToT with respect to the intermediate reasoning steps  (best results in bold font).}
\label{tab:evaluation_Agent_ToT_final}
\begin{adjustbox}{width=0.9\textwidth, center}
\begin{tabular}{l|c|c|c}
\hline
\textbf{Model} & \textbf{Syntactical Correctness (\%)} & \textbf{Functional Correctness (\%)} & \textbf{Code Similarity} \\
\hline
GPT-3.5 \textbf{+ ToT}    & \textbf{96.34} & 50.09 & 62.38 \\
GPT-3.5 \textbf{+ Agent}  & 78.4  & \textbf{52.15} & \textbf{87.15} \\
\hline
GPT-4 \textbf{+ ToT}      & 96.12 & 68.76 & 60.52 \\
GPT-4 \textbf{+ Agent}    & \textbf{99.2}  & \textbf{70.45} & \textbf{95.45} \\
\hline
GPT-4-turbo \textbf{+ ToT}      & 96.05 & 71.42 & 63.2 \\
GPT-4-turbo \textbf{+ Agent}    & \textbf{99.33}  & \textbf{74.57} & \textbf{94.39} \\
\hline
\end{tabular}
\end{adjustbox}
\end{table*}


\paragraph{\textbf{RQ1(b). How do the reasoning abilities of the Agent architecture compare against CoT and ToT for the program equivalence task?}}
The Agent architecture outperforms both CoT and ToT for the majority of the recorded metrics, with the biggest improvements obtained for functional correctness and code similarity. 
This implies that the intermediate programs generated by the Agent architecture and more likely to maintain the semantics of the source program.

For CoT, the code similarity metric of the final program improves on average with 41.1\% for GPT-3.5, 46.89\% for GPT-4, and 44.12\% for GPT-4-turbo. For ToT, the improvements of this metric are 24.77\% for GPT-3.5, 34.93\% for GPT-4, and 
31.19 for GPT-4-turbo.

This means that the final programs generated by the Agent architecture are much more similar to the target program than those generate by the CoT and ToT prompts. 
This supports our hypothesis that being able to explore different paths through the search space, as well as having access to domain specific information (i.e. syntax checker, unit tests, and code similarity metric) leads to better performance. 

%

As an additional remark, the Granularity column in Table \ref{tab:evaluation_Agent_CoT} suggests that the Agent architecture tends to take finer grained intermediate steps than CoT. It seems reasonable that smaller steps may result in better overall reasoning, and it is in line with out initial hypothesis.




\begin{tcolorbox}[colback=gray!7, colframe=gray!7, boxrule=0pt, left=5pt, right=5pt, top=5pt, bottom=5pt]
\textbf{RQ1 Answer}: When assessing the intermediate reasoning steps, both CoT and ToT prompting generate low-quality outputs. The resulting intermediate and final mutated programs frequently fail to maintain functional correctness, and the final mutated programs show reduced code similarity to the target program. This suggests possible  limitations with long-term proof planning for CoT and ToT prompting.
 In comparison, our Agent architecture surpasses both CoT and ToT, likely due to its non-linear reasoning abilities and access to domain-specific information when assessing candidate intermediate steps (i.e. programs). 
\end{tcolorbox}

\subsection{Evaluation of the downstream task}
\label{sec:downstream-evaluation}

In this section, we investigate the actual answers given to the equivalence queries in four configurations: (1) no intermediate reasoning provided, (2) CoT, (3) ToT, and (4) the Agent architecture. The results are given in Table~\ref{tab:evaluation_results_equivalence}.  
As metrics, we consider recall, precision, the F1 score and the area under the receiver operating characteristic (ROC AUC). For all the metrics, higher values correspond to higher performance of the model. 


\paragraph{\textbf{RQ2(a). How do CoT prompting and ToT prompting perform with respect to answering equivalence queries?}}

The results for no intermediate reasoning, CoT and ToT are all very close, with the order varying for different models (for GPT-3.5 and GPT-4, CoT and ToT are doing slightly better than no reasoning, whereas for GPT-4-turbo, no reasoning performs somewhat better).
This is surprising given that intermediate reasoning is generally expected to facilitate solving downstream tasks. One explanation might be that the intermediate reasoning elicited by CoT and ToT for this problem is too poor to actually help.




\paragraph{\textbf{RQ2(b). How does the Agent architecture answer equivalence queries compared to CoT and ToT prompting?}}
The Agent architecture outperforms all the other configurations suggesting that, when accurate,  intermediate reasoning does improve downstream task outcomes.

\begin{table*}[h]
\centering
\caption{Comparison between no intermediate reasoning, the Agent architecture, CoT, and ToT with respect to answering equivalence queries (best results in bold font).}
\label{tab:evaluation_results_equivalence}
\begin{adjustbox}{width=0.7\textwidth, center}
\begin{tabular}{l|c|c|c|c}
\hline
\textbf{Model} & \textbf{F1} & \textbf{ROC AUC} & \textbf{Precision} & \textbf{Recall} \\
\hline
GPT-3.5 {\bf + no reasoning} & 0.4969 & 0.4938 & 0.4938 & 0.5000 \\
GPT-3.5 {\bf + CoT} & 0.5644 & 0.5563 &  0.5542 & 0.5750 \\
GPT-3.5 {\bf + ToT} & 0.5542 & 0.5375 & 0.5349 & 0.5750 \\
GPT-3.5 \textbf{+ Agent} & \textbf{0.5839} & \textbf{0.5812} & \textbf{0.5802} & \textbf{0.5875} \\
\hline
GPT-4  {\bf + no reasoning} & 0.6135 & 0.6062 & 0.6024 & 0.6250 \\
GPT-4 {\bf + CoT} & 0.6234 & 0.6375 & 0.6486 & 0.6000 \\
GPT-4 {\bf + ToT} & 0.6194 & 0.6313 & 0.6400 & 0.6000 \\
GPT-4  \textbf{+ Agent} & \textbf{0.7248} & \textbf{0.7438} & \textbf{0.7826} & \textbf{0.6750} \\
\hline
GPT-4-turbo  {\bf + no reasoning} & 0.6829 & 0.6750 & 0.6667 & 0.7000 \\
GPT-4-turbo {\bf + CoT} & 0.6506 & 0.6375 & 0.6279 & 0.6750 \\
GPT-4-turbo {\bf + ToT} & 0.6135 & 0.6062 & 0.6024 & 0.6250 \\
GPT-4-turbo \textbf{+ Agent} & \textbf{0.7470} & \textbf{0.7375} & \textbf{0.7209} & \textbf{0.7750} \\
\hline
\end{tabular}
\end{adjustbox}
\end{table*}



\begin{tcolorbox}[colback=gray!7, colframe=gray!7, boxrule=0pt, left=5pt, right=5pt, top=5pt, bottom=5pt]
    \textbf{RQ2 Answer}: 
    When answering the program equivalence task, the Agent architecture outperforms all the other configurations (no intermediate reasoning, CoT prompting, ToT prompting), suggesting that better intermediate reasoning may lead to better outcomes for the downstream task.
\end{tcolorbox}

\subsection{Additional Agent evaluation}
\label{sec:agent-evaluation}

In this section, we further analyse of the Agent architecture to identify its most critical components and to assess whether a simpler architecture could achieve similar performance. We assess the Agent's performance with respect with its intermediate reasoning abilities, as this is the main objective of our architecture.
In order to limit cost (both dollar and compute), these experiments are only performed for GPT-3.5.

\paragraph{\textbf{RQ3(a). How does the Agent's backtracking ability affect its performance?}}

To answer this RQ, we perform an ablation study aimed at determining how much of the Agent's effectiveness is attributed to its ability to backtrack.
For this purpose, we retrain the model with backtracking disabled (while retaining the other possible actions). Results are given in Table~\ref{table:times-backtracking}. The Agent without backtracking abilities results in poorer metric values than the one with backtracking.

\begin{table}
\centering
\caption{Agent with and without backtracking}
\label{table:times-backtracking}
\begin{adjustbox}{width=0.59\textwidth}
\begin{tabular}{lcc}
\hline
                Metric &  w/o backtracking &  w backtracking \\
\hline
Syntactical Correctness  &                    76.64\% &                  {\bf 79.29\%} \\
Functional Correctness &                    57.93\% &                  {\bf 71.22\%} \\
   Code Similarity &                     53.20\% &                  {\bf 76.03\%} \\
\hline
\end{tabular}
\end{adjustbox}
\end{table}

\begin{figure}
    \centering
    \includegraphics[width=0.6\linewidth]{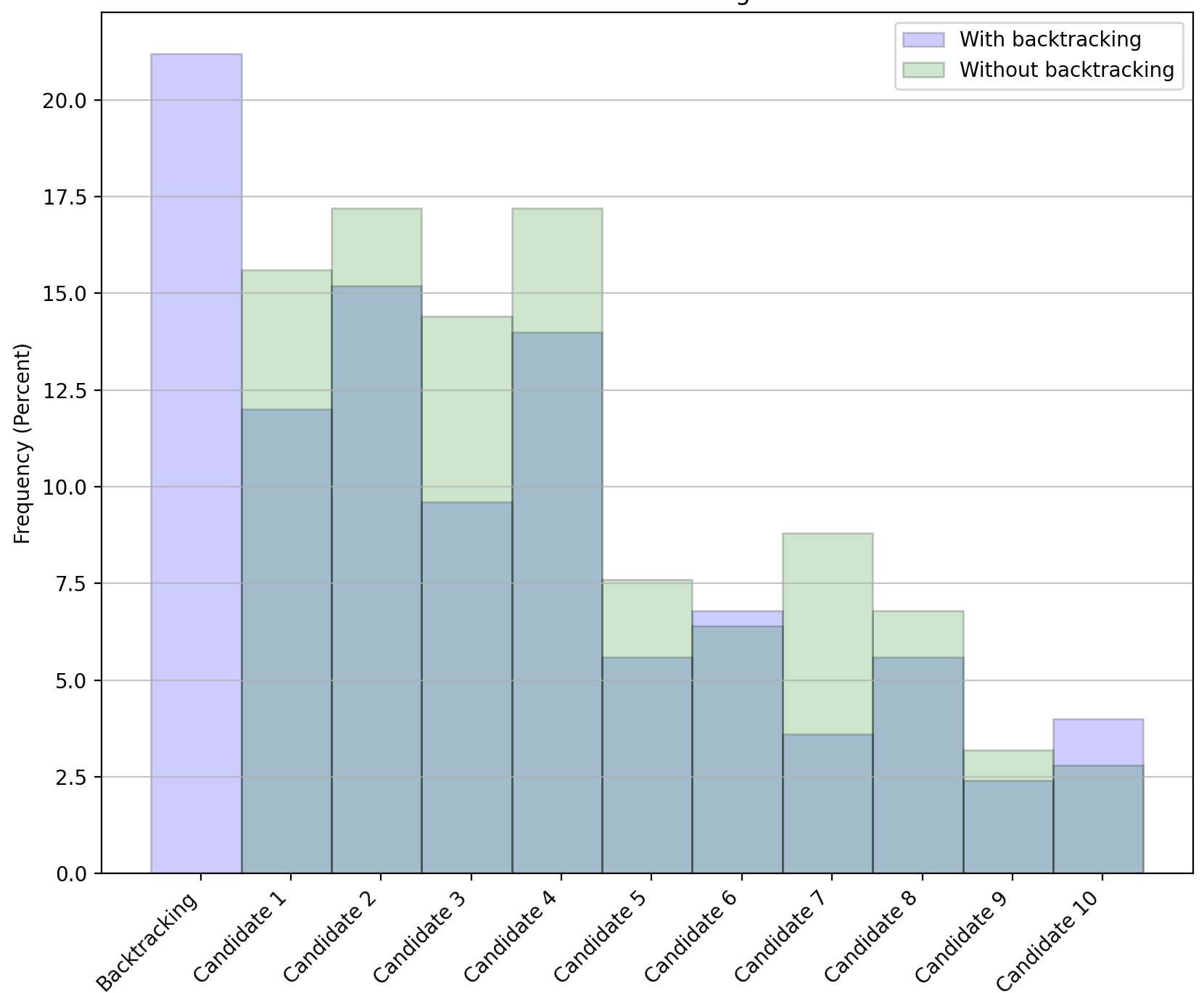}
    \caption{Histogram comparing the frequency distribution of discrete actions taken by two types of Agents: with and without backtracking. The initial action corresponds to backtracking (when available), whereas the other actions correspond to picking one of the candidate mutated programs. (Best viewed in color).}
    \label{fig:histo}
\end{figure}

Additionally, we investigate how often the agent chooses to backtrack. Figure \ref{fig:histo} denotes a histogram of the frequency of each action taken by the Agent, with and without backtracking. When backtracking was present,  it accounted for 21.13\% of actions, while the LLM's default decision (i.e., picking Candidate 1) was just selected in 13.45\% of the cases. Additionally, the data suggests a mild preference for picking candidates that are ranked higher by the LLM, visible by a slightly left-skewed distribution. The ability to backtrack appears to slightly reduce this preference.

\begin{tcolorbox}[colback=gray!7, colframe=gray!7, boxrule=0pt, left=5pt, right=5pt, top=5pt, bottom=5pt]
\textbf{RQ3(a) Answer}: Removing backtracking from our model significantly reduced its reasoning effectiveness, as shown by a reduction in all metrics in Table~\ref{table:times-backtracking}.
`\end{tcolorbox}

\paragraph{\textbf{RQ3(b). How does the Agent's learned policy compare against a greedy approach?}}

Rather than learning a policy, here we experiment with a few cheaper greedy approaches where space exploration is rule-based.
In particular, 
Policy 1 favors syntactically correct candidates, Policy 2 prioritizes functional correctness, 
and Policy 3 chooses the candidate with the highest sum of the normalised metrics for syntactical correctness, functional correctness, and code similarity to the target program.


Table \ref{tab:evaluation_manual_policies} presents a summary of the results, showing that the neural Agent outperforms all the greedy approaches (among which Policy 3 is the most effective), showing that training may be beneficial for the current problem.


\begin{table*}
\centering
\caption{Comparison between the Agent architecture and greedy approaches.
}
\label{tab:evaluation_manual_policies}
\begin{adjustbox}{width=1.0\textwidth, center}
\begin{tabular}{lcc|cc|c|c}
\hline
\textbf{Model} & \multicolumn{2}{c|}{\textbf{Syntactical Correctness (\%)}} & \multicolumn{2}{c|}{\textbf{Semantic Correctness (\%)}} & \multicolumn{1}{c|}{\textbf{Code Similarity}} & \textbf{Granularity} \\
 & mean & final & mean & final & final & mean \\
\hline
GPT-3.5 + Policy 1 & 67.43 & 62.93 & 51.28 & 48.10 
& 48.18 & 12.64 \\
GPT-3.5 + Policy 2 & 74.36 & 78.30 & 52.98 & 53.33 
& 57.40 & 32.40 \\
GPT-3.5 + Policy 3 & 76.54 & 73.38 & 66.09 & 64.36 
& 65.06 & 43.65 \\
GPT-3.5 \textbf{+ Agent} & {\bf 81.25} & {\bf 78.40} & {\bf 72.34} & {\bf 52.15} 
& {\bf 87.15} & {\bf 48.92} \\

\hline
\end{tabular}
\end{adjustbox}
\end{table*}

\begin{tcolorbox}[colback=gray!7, colframe=gray!7, boxrule=0pt, left=5pt, right=5pt, top=5pt, bottom=5pt]
 \textbf{RQ3(a) Answer}: The neural agent significantly outperforms all the greedy approaches.
\end{tcolorbox}

\paragraph{\textbf{RQ3(c) How does the Agent's performance vary when 
the source and the target programs use different programming concepts (e.g., data structures, control flow structures)?}}

The same programming task can be solved using different programming concepts (e.g., different data structures, different control flow structures). It's natural to presume that generating transformations between a source and a target program becomes more complex when the two programs make use of divergent strategies -- such programs correspond to Type-4 code clones \cite{DBLP:journals/tse/BellonKAKM07}. 

For illustration, below we present two distinct solutions for the Codeforces problem 255B: The programming task involves designing an algorithm that manipulates a string consisting of "x" and "y" by first swapping any adjacent "y" and "x" found starting at the start of the string, then removing any adjacent "x" and "y" pairs, repeatedly, until no more such operations can be applied. The algorithm prioritizes swapping over removal and processes the string from the beginning, ultimately printing the modified string as its result.
 
\begin{mdframed}[backgroundcolor=gray!5, linecolor=white]
\label{example:rq5-I}
\begin{lstlisting}[language=Python, caption={Solution: 2869941. This solution utilizes a dictionary to track the frequency of x and y in the string, allowing for a comparison of their counts to determine which character is more prevalent and by what margin.}]
count = { 'x': 0, 'y': 0 }
for i in raw_input():
    count[i] += 1

print(('x' if count['x'] > count['y'] else 'y') * abs(count['x'] - count['y']))
\end{lstlisting}

\begin{lstlisting}[language=Python, caption={Solution: 2793525. This solution employs a list as a stack to dynamically add or remove characters while iterating through the string, effectively canceling out adjacent 'xy' or 'yx' pairs. This stack approach enables a sequential evaluation and modification of the string, retaining only those characters that do not form cancelable pairs.}]
r = []
for x in raw_input():
	if not len(r) or r[-1] == x:
		r.append(x)
	else:
		r.pop()
print ''.join(r)
\end{lstlisting}
\end{mdframed}

We aim to investigate how the performance of the model with the Agent fluctuates when applied to programs using different programming concepts.
For this purpose, we randomly selected a subset of 143 Codeforces challenges from our dataset. From each challenge, we manually selected three Python solutions, A, B, and C, such that A and B use similar programming concepts, whereas C uses either a different data structure or a different control flow structure. Then, in our experiments, we compute program transformations for AB (A is the source program, and B is the target) and AC (A is the source program, and C is the target). 
 Again, tthis experiment was only performed for GPT-3.5.

\begin{table}[]
\centering
\caption{Comparison of source and target pairs implementing similar and different programming concepts}
\begin{adjustbox}{width=0.55\textwidth}
\begin{tabular}{lcccc}
\hline
\textbf{Metric}                             & \multicolumn{1}{c}{\textbf{Same} (AB)}  & \multicolumn{1}{c}{\textbf{Different} (AC)} \\ 
\multicolumn{1}{l}{}                       &  \multicolumn{1}{c}{final}   & \multicolumn{1}{c}{final}   \\ \hline
\multicolumn{1}{l}{Syntactical Correctness (\%)}                         & {\bf 94.48}                        & 89.39    \\ 
\multicolumn{1}{l}{Functional Correctness (\%)}                    & {\bf 82.75}                       & 69.12    \\ 
\multicolumn{1}{l}{Code Similarity}                            & {\bf 86.50}                       & 77.83   \\ \hline
\end{tabular}
\end{adjustbox}
\label{tab:sameDiff}
\end{table}

In Table \ref{tab:sameDiff}, we present a breakdown of the evaluation results for AB and AC.
We assess the metrics for 
the final transformed program. 
Overall, the analysis suggests that program pairs implementing similar 
programming concepts tend to achieve higher scores across all evaluated metrics compared to pairs using different concepts. For programs in the AC group, the syntactical correctness of the final program decreases by 5.09\% on average, functional correctness by 13.63\% on average, and code similarity to the target program by 8.67\% on average.
%

\begin{tcolorbox}[colback=gray!7, colframe=gray!7, boxrule=0pt, left=5pt, right=5pt, top=5pt, bottom=5pt]
 \textbf{RQ3(c) Answer}: Compared to programs using similar programming concepts, for programs implementing different concepts, syntactical correctness of the final program decreases by 5.09\% on average, functional correctness by 13.63\% on average, and code similarity to the target program by 8.67\% on average.
\end{tcolorbox}

\paragraph{\textbf{RQ3(d). What does the Agent learn?}}

\begin{figure}
    \centering
    \includegraphics[width=0.5\linewidth]{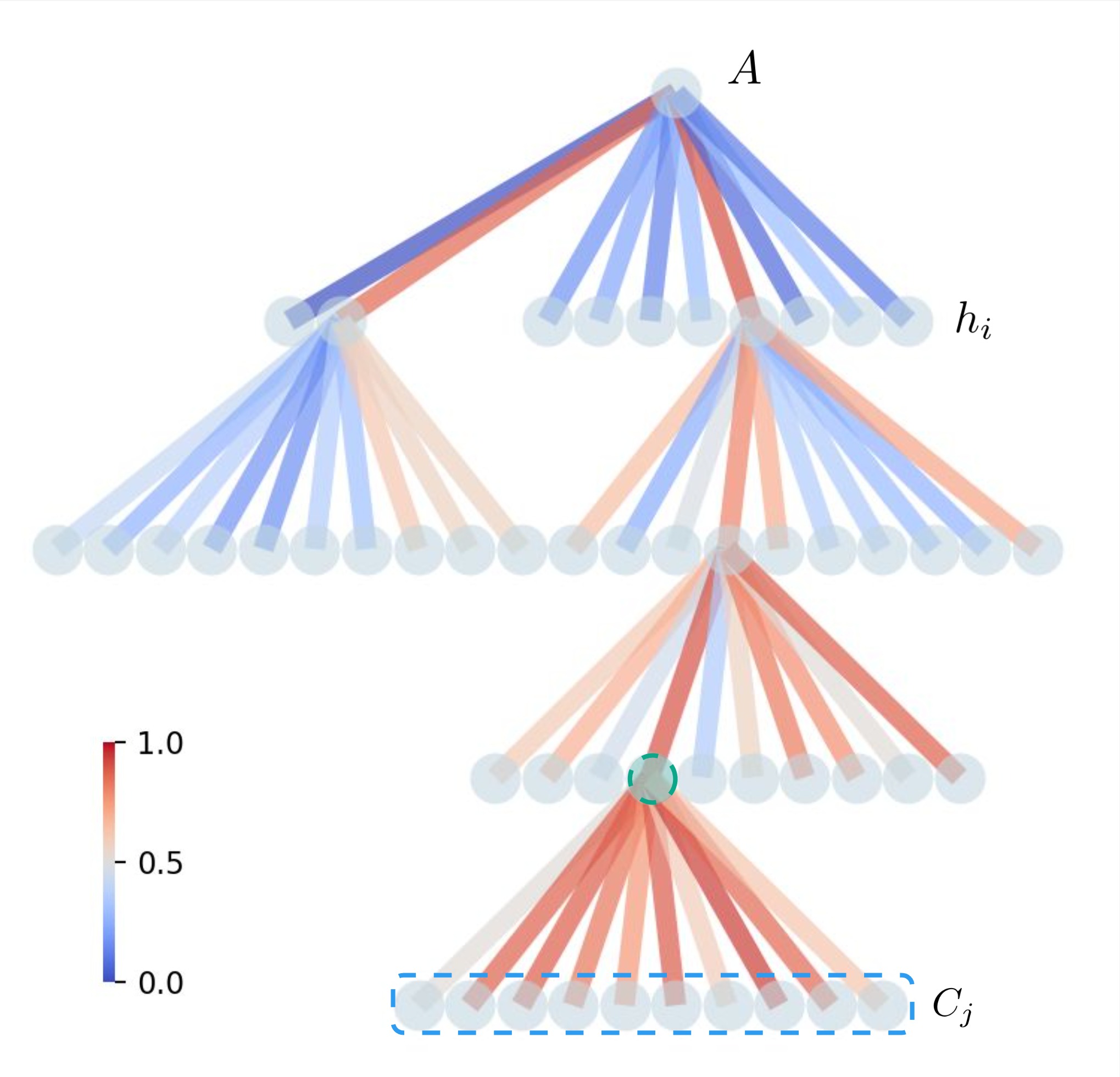}
    \caption{Visualization of a reasoning tree displaying graph attention on edges, where high attention scores are represented by red and low attention scores are depicted by blue edges. (Best viewed in color).}
    \label{fig:graphAttention}
\end{figure}

We randomly picked a reasoning tree and conducted an analysis of the attention scores obtained from the graph attention layers when the Agent is exploring candidate programs from the current node (the green node), as depicted in Figure \ref{fig:graphAttention}. We observed that recent candidates and candidates that have been selected in the past receive more attention. Also, it doesn't seem to be the case that the candidates top-ranked by the LLM always attract higher attention scores. 

In a separate analysis focused on the feature vectors, we investigated the influence weights for each feature in a vector $h_i$, and determined that the most important are syntactical and functional correctness, as well as similarity to the target program, which is very intuitive.

\begin{tcolorbox}[colback=gray!7, colframe=gray!7, boxrule=0pt, left=5pt, right=5pt, top=5pt, bottom=5pt]
\textbf{RQ6 Answer}: Our analysis found that recent candidates in the Reasoning Tree receive the most attention and that the policy prioritizes syntactical and functional correctness, as well as similarity to the target program.
\end{tcolorbox}

\section{Threats to validity}
\label{sec:threats-validity}

Our main result is that an RL agent trained with domain-specific information can help guide the LLM's space exploration, and thus produce better results when answering equivalence queries.  We discuss threats to the validity
of this conclusion. 

The number of benchmarks, as well as their length and number of tokens, could limit the generalisation of our findings. Moreover, we only experimented with a limited number of LLMs, and we can't guarantee that our results apply to others. We only applied the Agent architecture to the problem of answering equivalence queries. 
 Thus, we can't say whether the same improvement would be obtained for other tasks.
%
%
We judge functional correctness with respect to unit tests. However, unit tests 
may not fully capture a program's behaviour, which means that we may over-report functional correctness. 
%
%
Especially for CoT and ToT, different prompts may results in better results. To reduce the danger, we used prompts proposed in the literature. 

\subsubsection*{Future work}
We aim to explore the application of a reasoning tree that stores embeddings related to the program itself instead of solely relying on computed features from these embeddings. We also plan to apply our Agent architecture to other tasks.

\section{Related Work}
\label{sec:related-work}
\subsection{External reasoning and space exploration for LLMs.}

While, initially, LLMs were used as black-box, monolithic entities, recently, there has been a shift towards architectures that foster some form of logical reasoning as part of the problem-solving process, sometimes by leveraging additional, possibly non-neural systems. For instance, Karpas et al. propose a neuro-symbolic architecture, dubbed the Modular Reasoning, Knowledge, and Language system, denoting a flexible architecture with multiple neural models, complemented by discrete knowledge and reasoning modules \cite{DBLP:journals/corr/abs-2205-00445}. Lazaridou et al. embed the model within a simple program which, during inference, allows the system to leverage results returned from the web using Google Search without the need for a dedicated retrieval-augmented model. The Chain of Thought approach aims to improve the ability of LLMs to perform complex reasoning by generating a series of intermediate reasoning steps  \cite{DBLP:conf/nips/Wei0SBIXCLZ22}.

Other works are focused on decision-making and space exploration, given that LLMs were shown to have difficulty with proof planning when using a linear search strategy. In particular, when multiple valid deduction steps are available, they are not able to systematically explore the different options \cite{DBLP:conf/iclr/Saparov023}. Several works attempt to ameliorate this. Schlag et al. introduced LLM programs, where an LLM is embedded in a classic program to carry out more complex tasks ~\cite{DBLP:journals/corr/abs-2305-05364}. Then, this method decomposes the main problem recursively into subproblems until they can be solved by a single query to the model. LLM+P \cite{DBLP:journals/corr/abs-2304-11477} delegates the actual planning process to a classical planner in order to solve long-horizon robot planning problems. Tree of Thoughts integrates thought sampling and value feedback, enabling effective search inside a reasoning tree \cite{DBLP:conf/nips/YaoYZS00N23}. It implicitly builds in decision-making and planning. A similar architecture is proposed by Long, who augments an LLM with additional modules, including a prompter Agent, a checker module, a memory module, and a ToT controller. 

The approaches based on ToT are the closest to our work. However, there are several differences. While \cite{DBLP:conf/nips/YaoYZS00N23} uses DFS/BFS/beam search, we make use of an RL Agent. Instead of using the LLM to evaluate the potential of different candidates,  the Agent has access to domain-specific knowledge, which in our case are a syntax checker, unit tests, codeBLEU and Jaccard metrics. Conversely, from \cite{DBLP:journals/corr/abs-2305-08291}, we propose an architecture with a different way of exploring the reasoning tree, which only stores the feature vectors. Moreover, none of the existing works investigates the ToT space exploration for code related tasks.

\subsection{Reinforcement Learning and LLMs.}

In recent years, RL has become a common paradigm for fine-tuning LLMs. Many models are fine-tuned with RL, e.g. OpenAI's GPT-4~\cite{GPT4}, Google's Gemini \cite{gemini}, and Anthropic's Claude 3 \cite{claude3}.
While such fine-tuning generally makes use of Proximal Policy Optimization, in \cite{DBLP:journals/corr/abs-2306-11816}, Chang et al. focus on more efficient RL algorithms for fine-tuning LLMs on downstream tasks with predefined rewards (e.g. Bleu~\cite{papineni-etal-2002-bleu}, or reward learned from human preference
feedback). 
As opposed to focusing on the use of RL during training, CodeRL \cite{CodeRL} uses it to integrate a critic network that evaluates the functional correctness of the generated programs, enabling the refinement and repair of output programs based on their functional correctness during test time. 
Conversely to these works, we use RL to train an agent that drives the exploration of the solution space during LLM-based code generation.

\subsection{Program equivalence.}

Program equivalence has been studied for a long time. While many of the approaches are based on symbolic reasoning~\cite{DBLP:journals/dagstuhl-reports/LahiriMSU18,DBLP:journals/stvr/GodlinS13}, neural approaches have also emerged in recent years.
For instance, Kommrusch et al. propose a neural network architecture based on a transformer model to generate proofs of equivalence (expressed as a sequence of rewrites) between program pairs~\cite{Kommrusch2021c}. As opposed to our work, they consider a very conservative set of rewrite rules that are guaranteed to be semantics-preserving, and they only target straight-line programs. 
The authors previously applied a similar architecture to show equivalence between dataflow graphs with a graph-to-sequence neural model~\cite{Kommrusch2020}. The approach aims to automate the verification of dataflow graph equivalence by transforming the graphs into sequences and leveraging neural networks for classification.
They also presented a method for proving equivalence between complex mathematical expressions \cite{Kommrusch2021} using graph-to-sequence neural models. The approach involves representing expressions as graphs and transforming them into sequences, which are then processed by neural networks to classify their equivalence. 

A very related software engineering task is code clone detection, i.e. detecting duplicate or similar code. While many techniques and tools have been proposed over the years for code clone detection (see surveys \cite{ROY2009470,10.1007/978-981-15-7533-4_27}), they might work better for clones with high degree of syntactical similarity (Type-1, Type-2, and Type-3 with high syntactical similarity), but have weaker performance for clones with lower syntactical similarity (Type-3 and Type-4) \cite{DBLP:conf/icsm/SvajlenkoR15}.
More recently, neural approaches have been proposed for the related problem of code clone detection (see survey~\cite{DBLP:journals/jss/LeiLLAK22}). Generally, these work leverage neural networks to generate a vector
representation for each code fragment and then detect clones by computing the similarities between the vector representations of
two code fragments.
%
%
%
Dou et al. \cite{CodeCloneLLM} examines the effectiveness of LLMs in identifying code clones, 
and reveals that their performance improves with CoT prompting and vector embeddings. 
As opposed to this work, we are not focusing on code clone detection but rather investigating the exploration of the search space in the context of program transformation and program equivalence. Instead of directly outputting a classification judgment of equivalence, we investigate the ability of the LLM to generate a sequence of rewrites, transforming a source program into the target. 

\section{Conclusions}

We proposed an architecture where an
RL agent trained with domain-specific information (i.e., access to a syntax checker, a unit tests checker, and a code similarity checker) can help guide the LLM's space exploration in a non-linear manner. 
For the task of program equivalence, we compared the reasoning elicited by our architecture against CoT and ToT style prompts, as well as a greedy approach, and reported positive results. 

\section{Data Availability}
A replication package is available at \url{https://anonymous.4open.science/r/LLM-Agent-0307/README.md}.

\bibliographystyle{ACM-Reference-Format}
\bibliography{acmart}

\end{document}